\documentclass[letterpaper, 10 pt, conference]{ieeeconf}  

\IEEEoverridecommandlockouts                              

\let\labelindent\relax
\usepackage{mathptmx} 
\usepackage{times} 
\usepackage{amsmath} 
\usepackage{amssymb}  
\usepackage{graphicx}
\usepackage{subcaption}
\usepackage{caption}
\usepackage{booktabs}
\usepackage{multirow}
\usepackage{subcaption}
\usepackage{xcolor}
\usepackage{amsmath}
\usepackage{amssymb}
\usepackage{enumitem}
\usepackage{xcolor}
\usepackage{array}
\newcolumntype{P}[1]{>{\centering\arraybackslash}p{#1}}

\DeclareMathOperator*{\argmin}{arg\,min}
\usepackage[
  separate-uncertainty = true,
  multi-part-units = repeat
]{siunitx}
\title{\LARGE \bf
Deep Learning of Movement Intent and Reaction Time for EEG-informed Adaptation of Rehabilitation Robots}

\author{Neelesh Kumar and Konstantinos P. Michmizos, Member \textit{IEEE}
\thanks{*This work is supported through Grant K12HD093427 from the National Center for Medical Rehabilitation Research, NIH/NICHD.}
\thanks{NK and KM are with the Computational Brain Lab, Department of Computer Science, Rutgers University, New Jersey, USA. 
        {\tt\small michmizos@cs.rutgers.edu}}%
}
\begin{document}

\maketitle
\thispagestyle{empty}
\pagestyle{empty}

\begin{abstract}
Mounting evidence suggests that adaptation is a crucial mechanism for rehabilitation robots in promoting motor learning. Yet, it is commonly based on robot-derived movement kinematics, which is a rather subjective measurement of performance, especially in the presence of a sensorimotor impairment. Here, we propose a deep convolutional neural network (CNN) that uses electroencephalography (EEG) as an objective measurement of two kinematics components that are typically used to assess motor learning and thereby adaptation: i) the intent to initiate a goal-directed movement, and ii) the reaction time (RT) of that movement. We evaluated our CNN on data acquired from an in-house experiment where 13 subjects moved a rehabilitation robotic arm in four directions on a plane, in response to visual stimuli. Our CNN achieved average test accuracies of 80.08\% and 79.82\% in a binary classification of the intent (intent vs. no intent) and RT (slow vs. fast), respectively. Our results demonstrate how individual movement components implicated in distinct types of motor learning can be predicted from synchronized EEG data acquired before the start of the movement. Our approach can, therefore, inform robotic adaptation in real-time and has the potential to further improve one's ability to perform the rehabilitation task.
\end{abstract}
\section{INTRODUCTION}
By delivering intensive, challenging and task-specific training, rehabilitation robots~\cite{krebs1998robot} have now established themselves as an alternative treatment for sensorimotor impairments~\cite{winstein2016guidelines}. Given the importance of active participation during therapy for motor recovery~\cite{lotze2003motor}, there is an increasing interest in adapting the robotic therapy and tailoring it to the needs and skills of the patients~\cite{krebs2003rehabilitation,patton2004robot,patton2006evaluation}. The current 'assist-as-needed' adaptive controllers use movement-related kinematics feedback to alter their control strategies, and subsequently adapt the rehabilitation task components~\cite{krebs2003rehabilitation,marchal2009review}. Despite their demonstrated success in functional motor recovery~\cite{krebs2009working,michmizos2015robot}, the robot-derived movement kinematics is a subjective measurement of movement performance as it often becomes masked by the targeted sensorimotor impairments. This evidence-based adaptation~\cite{marchal2009review} limits our ability to study and improve the robotic intervention in rehabilitation.
\par
To overcome this limitation, a fascinating possibility is to have an adaptive strategy targeting specific movement components that are implicated in motor learning, either explicitly or implicitly- depending upon whether or not the patients deliberately aim to improve these attributes . One such component implicated in explicit motor learning is the intent of patients to initiate a movement. It has long been known that when rehabilitation training consists of voluntary movements, a significant improvement in the motor performance is achieved. This has been associated with the cortical reorganization of the motor cortex upon motor training~\cite{lotze2003motor}. In our previous work, we used the initiation of voluntary movements as a faithful indicator of explicit motor learning~\cite{michmizos2015robot}, followed by clinical outcome~\cite{michmizos2017pediatric}. Another movement component, implicated in implicit motor learning, is the reaction time (RT) of a movement, defined as the time passed between the onset of a stimulus and the start of the movement~\cite{light1996reaction,michmizos2014reaction}. We have previously shown that there is a decrease in average RT of the patients as the training progresses, further confirming its role as a metric for motor learning that is used to assess motor deficits~\cite{rogers1988motor,marsden1982mysterious} and improvement~\cite{michmizos2015robot}. Given their immediate applicability in assessing both implicit and explicit motor learning, these two movement components can be used as adaptation criteria. 
\par
For the adaptation to become clinically important, one needs to first define the criteria that are indicative of the motor performance and then to objectively measure them. Arguably, the most faithful representation of movement is in its underlying neuronal activity. 
Indeed, for more than half a century, a significant amount of studies have identified specialized neurons in the motor cortex and other cortical and sub-cortical areas that encode movement components, including direction~\cite{georgopoulos1986neuronal}, amplitude~\cite{pruszynski2008rapid}, force~\cite{evarts1968relation} and speed~\cite{tankus2009encoding}. Interestingly enough, a recent study showed that the motor thalamic neurons encode the movement intent and RT~\cite{gaidica2018distinct}. Overall, neuronal activity offers an objective assessment of movement components in healthy and impaired subjects.
\par 
By virtue of their non-invasiveness and high temporal resolution, electroencephalography (EEG) has established itself as an objective measure for assessing motor behavior~\cite{tacchino2016eeg,jin2006alpha}, and has also become a reliable measurement of the cognitive functions involved in the execution of motor tasks in stroke~\cite{muralidharan2011extracting,park2016eeg}. The first approaches to decode EEG relied on hand-crafted features such as common spatial pattern~\cite{ang2008filter} and classifiers such as support vector machines (SVM) to segregate those features. However, these methods do not generalize well to new subjects~\cite{lotte2010regularizing}, and their need for hand-crafted features further impedes their use in real-time. Deep neural networks~\cite{lecun2015deep} overcome these limitations through unsupervised feature extraction and powerful generalization~\cite{zhang2016understanding}. These networks have recently started to be used in classifying EEG signals for various mental decoding tasks with remarkable performance \cite{schirrmeister2017deep}.
\par
Here, we propose a deep learning approach to predict two fundamental movement components associated with motor learning, namely the intent to initiate a goal-directed movement and RT, using EEG data as a measurement that circumvents the limitations imposed by the impaired kinematics. To do so, we developed a convolutional neural network (CNN) that was trained to decode EEG activity and predict: i) the intent of the subject to initiate a movement (intent vs. no intent), and ii) RT in initiating the movement (2 classes- fast vs. slow). We evaluated our CNN on data acquired from an in-house experiment where 13 subjects moved a rehabilitation robotic arm in four directions on a plane in response to visual stimuli. The CNN achieved average test accuracies of 80.02\% and 79.82\% for movement intent and RT classification respectively. These results support our ongoing effort to devise a rehabilitation strategy that adapts based on an objective measurement of specific movement components, further personalizing therapy.

\section{Methods}
\subsection{Subjects and Experimental Apparatus}
\begin{figure}[t]
\vspace{5.2pt}
    \centering
    \includegraphics[width=8.5cm]{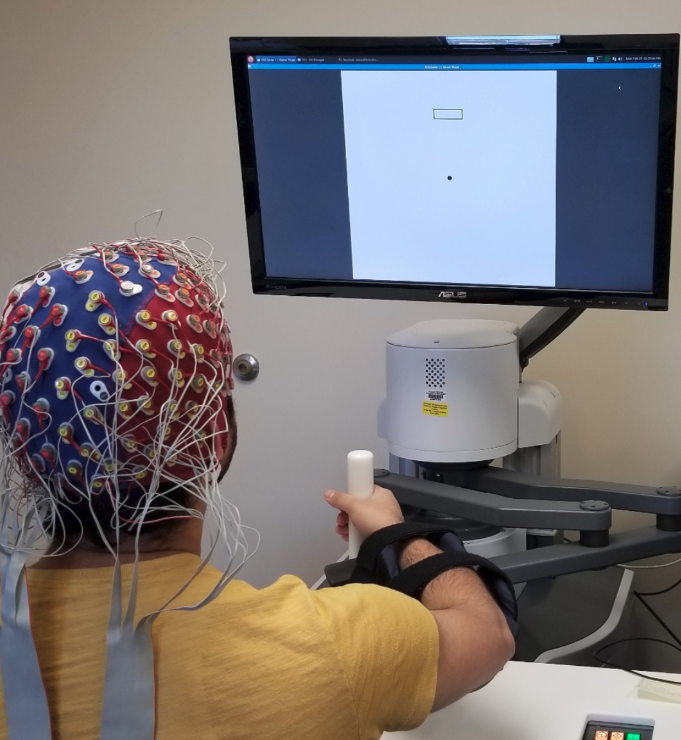}
    \caption{Experimental Setup: Subjects executed a goal-directed motion task in their interaction with the Bionik InMotion Arm rehabilitation robot. EEG data were acquired as subjects performed the task in synchrony with the kinematics recordings recorded by the robot.}
    \label{fig:exp}
\end{figure}

Thirteen healthy subjects (age=23$\pm$2, 5 females, right handed) participated in this experiment. Subjects provided written consent and the experimental protocol was approved by the local Institutional Review Board (IRB). The EEG data were recorded using a 128-channel Biosemi ActiveOne EEG system with sampling frequency of 1024 Hz. The motor task was performed on the InMotion Arm Robot (Bionik Laboratories Corp.)(Figure \ref{fig:exp}), with the robot data collected at a sampling frequency of 180 Hz. Subjects were seated at an appropriate distance from the screen so that they could perform the task comfortably without moving their torso. 
\begin{figure*}
\vspace{5.2pt}
    \includegraphics[]{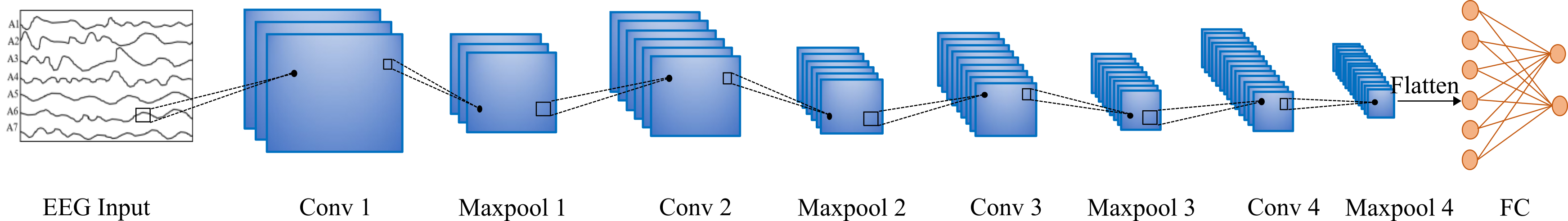}
    \caption{The 5 layer CNN architecture. The inputs to the CNN were EEG data of dimension 128 x 125 (0.5s recordings from 128 channels sampled at 250Hz). The filter size for convolution was 5 x 5 for movement intent and 3 x 5 for RT for all the layers. Batch normalization and Maxpooling was applied to the output of each convolutional layer. The output of the convolution layer was flattened and the resulting feature vector was passed through a fully connected layer. Finally, softmax was applied to convert the output of fully connected layer to class probabilities.}
    \label{fig:architecture}
\end{figure*}

\subsection{Experiment}
Building on a previous experimental protocol \cite{michmizos2014frontiers, michmizos2014reaction}, we developed a goal-directed task and asked the subjects to perform it on the arm rehabilitation robot. The task environment comprised of a pointer and a target box (Figure \ref{fig:exp}). The pointer indicated the current position of the end-effector of the robotic arm in the 2D plane of motion. Subjects were asked to move the pointer to the target box ``as fast as possible''. After the pointer entered the target box, the next target box appeared with a delay of 20ms with small added jitter. The target appeared at random in any of the four orthogonal directions- left, right, up or down. Subjects performed the task in two modes: i) active mode, where subjects performed the motion, and ii) passive mode, where motion was performed by the robot with the subject's arm affixed to the robotic end effector. Each subject performed 210 active and 210 passive motion trials. 

\subsection{EEG Minimal Preprocessing and Labeling}
Minimal data preprocessing was done to get rid of the contamination in the EEG data, as described below. The minimal preprocessing on the one hand ensured the CNN would learn the representations by itself and, on the other hand,  made it possible for the CNN to be compatible with a real-time system. A bandpass filter of 1 Hz - 40 Hz was applied to remove the low and high frequency artifacts and drifts. Independent component analysis (ICA) was used to get rid of occular artifacts. We then segmented the data into trials containing the events of interest. The time window for segmentation was chosen by grid search: -0.5s to 0.0s for both movement intent classification and RT (with t=0s indicating the start of the motion). We normalized the segmented trials using z-score normalization and downsampled all the trials to 250 Hz to reduce the computational load.  For classifying the movement intent, all active motion trials were labelled as 'Intent' and all passive motion trials were labelled as 'No Intent'. The movement intent dataset had 210 data points for each class per subject. To compute the RT, we measured the time difference between the onset of a stimulus and the start of the movement, where the movement was said to be started when the velocity exceeded a certain threshold. We discretized the RT into two classes- fast and slow, by choosing suitable thresholds that were determined from the histogram of RTs for each subject separately, based on the distribution of their RT across the experiement. In addition, all trials corresponding to RTs that were outliers, i.e. less that 0.15 seconds and greater than 0.8 seconds were removed from consideration, similarly to our previous studies \cite{michmizos2014reaction,michmizos2014modeling}.
This resulted in RT datasets of varying size for each subject, with an average size of 75 data points per class (SD = 14). 

\subsection{Notations}
For each subject $i$, we created a dataset $D^i = \{(X_i^1, y_i^1), (X_i^2, y_i^2),.....,(X_i^{n_i}, y_i^{n_i})\}$, where $n_i$ denotes the number of trials recorded for that subject. For every trial $j$, $X^j \in \mathbb{R}^{128 \times T}$ is a 2 dimensional matrix which was created by stacking recordings from all the channels. $T$ is the duration of recording of each trial. The labels $y^j$ of the trial $j$ contained a value from $\{0,1\}$ corresponding to the two classes. 

\subsection{The Convolutional Neural Network}
We developed a multi-layered CNN (Figure \ref{fig:architecture}) for classifying movement intent and RT, that performed convolution in spatial and temporal space. The inputs to both networks were preprocessed EEG trials acquired from 128 channels and sampled at 250 Hz. We passed the outputs of each convolutional layer through ReLU non-linearities and then applied batch normalization, to normalize the ReLU outputs to zero mean and unit variance. Batch normalization has regularization properties and is helpful in preventing overfitting~\cite{ioffe2015batch}. We also applied max pooling at the end of each layer to reduce computational load. Max pooling has also the desirable property of translational invariance which often gives better generalization across subjects. The last layer in both networks was a fully connected layer with softmax that took in the flattened feature vector produced by the last convolutional layer and converted it to class probabilities.  The choice of the CNN hyper-parameters, i.e. the number of layers, kernel size, etc. were limited by the training data size and the input dimension, and were found using a grid search over the allowable hyper-parameters space. The CNN architectures for classifying movement intent and RT are shown in Tables \ref{tab:archact} and \ref{tab:archrt}. 
\begin{table}[h]
\normalsize
  \begin{center}
    \caption{CNN Architecture for Movement Intent}
    \label{tab:archact}
    \begin{tabular}{|P{0.4\linewidth}|P{0.4\linewidth}|}
      \hline 
      \textbf{Layers} & \textbf{Output Shape}\\
      \hline 
      Conv1 (5x5)  & 32 x 124 x 121 \\
      ReLU + BatchNorm  & 32 x 124 x 121 \\
      MaxPool (3x3)  & 32 x 41 x 40 \\
      Conv2(5x5)  & 64 x 37 x 36 \\
      ReLU + BatchNorm  & 64 x 37 x 36 \\
      MaxPool (3x3)  & 64 x 12 x 12 \\
      Conv3 (5x5)  & 128 x 8 x 8 \\
      ReLU + BatchNorm  & 128 x 8 x 8 \\
      MaxPool (3x3)  & 128 x 2 x 2 \\
      Fully Connected  & 1 x 2 \\
      Softmax & 1 x 2 \\
      \hline 
    \end{tabular}
  \end{center}
\end{table}

\begin{table}[h]
\normalsize
  \begin{center}
    \caption{CNN Architecture for RT}
    \label{tab:archrt}
    \begin{tabular}{|P{0.4\linewidth}|P{0.4\linewidth}|}
      \hline 
      \textbf{Layers} & \textbf{Output Shape}\\
      \hline 
      Conv1 (3x5)  & 32 x 126 x 121 \\
      ReLU + BatchNorm  & 32 x 126 x 121 \\
      MaxPool (2x2)  & 32 x 63 x 60 \\
      Conv2(3x5)  & 64 x 61 x 56 \\
      ReLU + BatchNorm  & 64 x 61 x 56 \\
      MaxPool (2x2)  & 64 x 30 x 28 \\
      Conv3 (3x5)  & 128 x 28 x 24 \\
      ReLU + BatchNorm  & 128 x 28 x 24 \\
      MaxPool (2x2)  & 128 x 14 x 12 \\
      Conv4 (3x5)  & 256 x 12 x 8 \\
      ReLU + BatchNorm  & 256 x 12 x 8 \\
      MaxPool (2x2)  & 256 x 6 x 4 \\
      Fully Connected  & 1 x 2 \\
      Softmax & 1 x 2 \\
      \hline 
    \end{tabular}
  \end{center}
\end{table}

\subsection{Network Training}
The CNN computed a mapping from the EEG trial to the labels, $f(X^j, \theta):\mathbb{R}^{128 \times   T}\to\mathbb\{0,1\}$ where $\theta$ were the trainable parameters of the network. The network was trained to minimize the average loss over all training examples:
\begin{equation}
    \hat{\theta} = \argmin \frac{1}{N} \Sigma_{i=1}^N l(X^i, y^i;\theta),
\end{equation}
where $N$ denotes the number of training examples and $l$ is the loss function, which in our case was the binary cross entropy loss function. The batch size was 64.  For optimization, we used Adam, a variant of stochastic gradient descent, with learning rate of $10^{-3}$.  
\begin{figure*}[t!]
\vspace{5.2pt}
    \centering
    \includegraphics[]{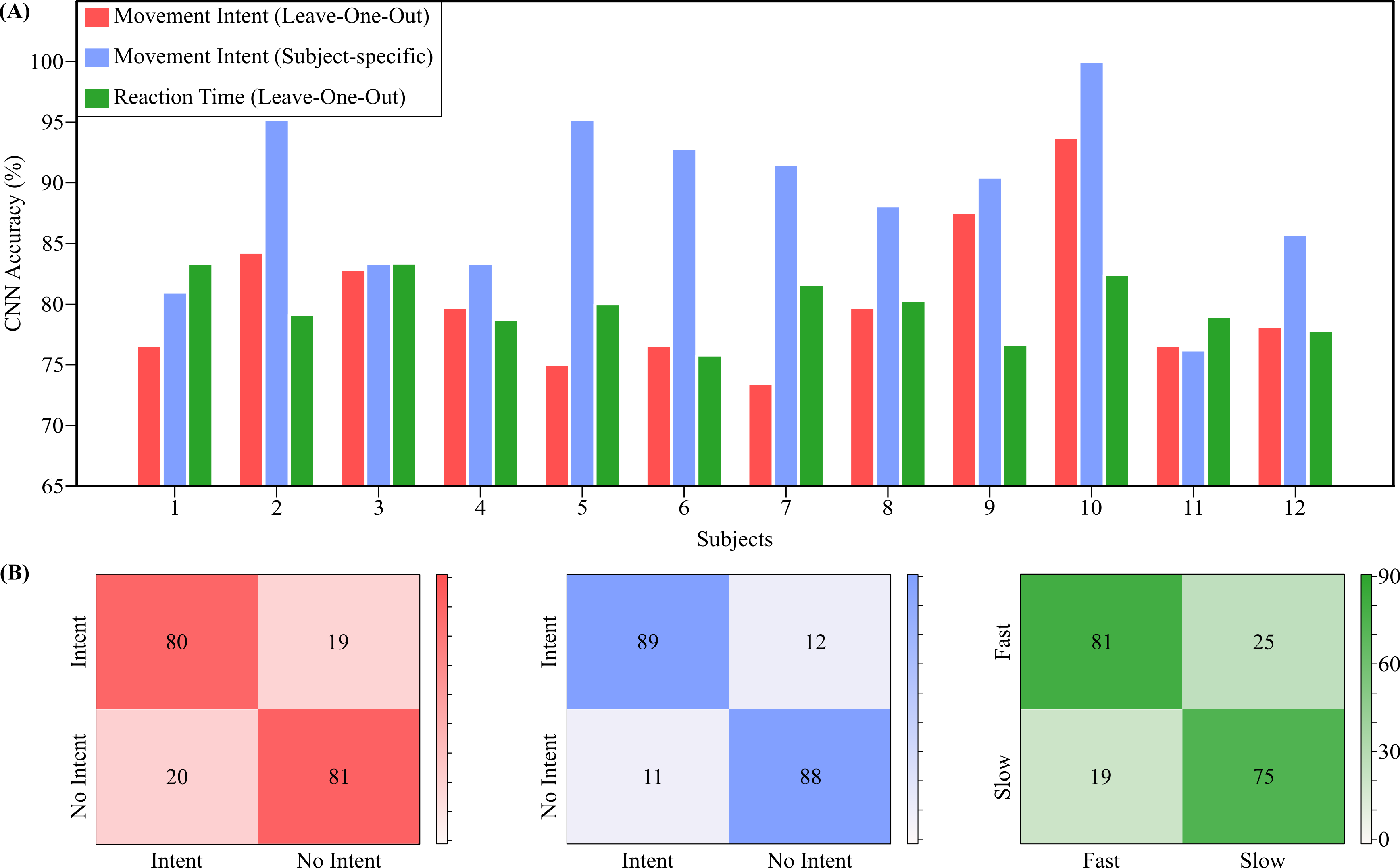}
    \caption{\textbf{A}: CNN accuracies for the 2 classification tasks when evaluated using leave-one-out and subject-specific training. \textbf{B}: Left - Averaged confusion matrix for movement intent classification evaluated using leave-one-out; Center - Averaged confusion matrix for movement intent classification evaluated using subject-specific training; Right - Averaged confusion matrix for RT classification evaluated using leave-one-out. Numbers are in percentage.}
    \label{fig:results}
\end{figure*}

\subsection{Network Validation}
We evaluated our proposed CNN in three ways: 
\begin{enumerate}
    \item Leave-one-out: Data from all but one subject were used for training. Evaluation was done on the left-out subject. This tested the CNN's ability to generalize to new subjects, that were not included in training.
    \item Subject-specific training: Data from a single subject were split randomly into training and test in the ratio 4:1. Validation was done on the test data. This tested the ability of the CNN to predict movement components when trained on individual subjects.
    \item All data: Data from all subjects were split randomly into training and test in the ratio 4:1. This allowed us to determine how well the CNN performs if it has access to data from a large number of subjects.
\end{enumerate}
While the movement intent classification was evaluated using all three evaluation techniques, RT classification was evaluated using only 1) and 3) since the number of data points per subject for RT were low to train a deep network, for the given variability of the labels (RT).\ 
\section{Results}
\subsection{Leave-one-out Evaluation}
To measure the generalization performance of our CNN, we evaluated it using the leave-one-out technique. Our CNN achieved an average accuracy of $80.08\% \pm 5.70\%$ for movement intent classification, and $79.82\% \pm 2.37\%$ for RT. Subject-specific results are shown in (Figure \ref{fig:results}A: red and green bars). While the CNN's prediction of movement intent was most accurate for subject 10 (93.75\%), the RT CNN had the best performance for subject 3 (83.34\%). In general, the CNN performed well enough for all subjects.

\subsection{Subject-specific Training}
We trained and evaluated our CNN for each subject separately, to test its ability to infer each individual's intent when data from only that person were available (Figure \ref{fig:results}A: blue bar). In this case, our CNN had an average accuracy of $88.97\% \pm 6.27\%$ for movement intent. Given that the network was trained for each subject separately, its increase in accuracy over leave-one-out was expected because of the reduced variability in the EEG.  

\subsection{Training on All Data}
Next, we evaluated our CNN on data from all the subjects, partitioned randomly into train and test in the ratio 4:1. This is a good indication of how our model can perform when data from a large pool of patients is available before-hand. The results, averaged over 10 random partitions, are shown in Table \ref{tab:eval2}. We notice that there is an improvement in the performance of the CNNs, when compared to leave-one-out evaluation, since the CNNs were trained on a more representative set of trials that were collected from all the subjects participated in the experiment.

\begin{table}
\vspace{5.2pt}
\normalsize
  \begin{center}
    \caption{Classification Accuracies for All data Training}
    \label{tab:eval2}
    \begin{tabular}{|P{0.4\linewidth}|P{0.4\linewidth}|}
      \hline 
      \textbf{Task} & \textbf{Mean Accuracy}\\
      \hline 
      Movement Intent & 87.34\% $\pm$ 2.83\% \\
      Reaction Time & 84.68\% $\pm$ 3.68\%  \\
      \hline 
    \end{tabular}
  \end{center}
\end{table}

\subsection{Confusion Matrices}
The averaged confusion matrices shown in Figure \ref{fig:results}B demonstrate that the network can predict both classes of the two classification tasks equally well. This is further revealed in the F1 scores, defined as the harmonic mean of precision and recall. Specifically, the F1 scores were 0.80 (leave-one-out) and 0.88 (subject-specific) for classifying the movement intent, and 0.80 for classifying RT (leave-one-out). 

\section{Discussion}
Here, we presented a deep learning framework to predict two movement components implicated in explicit and implicit motor learning, which are applicable in adapting rehabilitation robots: i) movement intent and ii) movement RT. Our framework consists of a CNN that associates the EEG activity with these components. The CNN achieved high test accuracies for all the evaluation cases, thus demonstrating its practical effectiveness in our ongoing effort to develop an adaptive robot based on individual components of movement. 
\par 
Here, we targeted the movement intent from the EEG activity because of the role of voluntary movements in improving motor performance~\cite{lotze2003motor}. Neurological deficits, particularly in the supplementary motor area (SMA), have been found to directly affect the ability of patients to initiate movement~\cite{krainik2001role}. Further,~\cite{krainik2004role} suggested a possible compensatory recovery mechanism through recruitment of healthy medial and lateral premotor circuitry. In the clinical setting, ours~\cite{michmizos2015robot} and other's~\cite{reinkensmeyer2014understanding} studies have revealed significant improvements in the number of self-initiated movements executed by the patients between admission and discharge. 
\par
We also targeted the prediction of RT from the EEG, because RT is one of the most well-studied behavioral indicators of neurological integrity. Interestingly, RT has not only been found to be related to neurological deficits, but also it is responsive to intervention~\cite{mirabella2013stimulation}, including exercise and practice~\cite{light1996reaction,marchal2009review}. In light of the fact that the sensorimotor control of the upper extremities is driven by cortical and subcortical areas, our finding that RT behavior is similar in lower and upper extremities \cite{michmizos2014reaction,michmizos2014modeling} expands the role of a presumptive supraspinal pathway from coordinating to controlling discrete lower extremity movements. In fact, when we incorporated RT in the clinical evaluation of the MIT-pediAnklebot ~\cite{michmizos2015robot}, we found statistically significant improvements in RT between admission and discharge, when children with cerebral palsy received therapy for 3 weeks. 
\par
Taken collectively, the EEG-derived movement components are promising in complementing the kinematics-based adaptation, further assessing and treating the neurological impairments. A hybrid set of kinematics- and EEG-based measures of motor learning could help robots to adapt to one's cognitive and motor efforts that will no longer be masked by the inherent variability of sensorimotor impairments. For example, we have previously shown how cognitive components of a serious game can be adapted to promote patient engagement during sensorimotor therapy~\cite{kommalapati2016virtual}. In addition to opening up the possibilities of adapting gamification of rehabilitation when only gross kinematics measurements can be acquired, a number of robots have also been developed that encourage voluntary movements by remaining passive unless the patient initiates the movement~\cite{marchal2009review}. Further, by giving new knowledge on how EEG-features are related to distinct movement components, our deep learning approach can also be used to assess the level of neurological deficit, and how well such deficit responds to a particular treatment.
\par
Adding to the mounting studies demonstrating the effectiveness of deep neural networks in predicting cognitive functions from EEG~\cite{craik2019deep}, this paper shows how CNN can use its two-dimensional structure to extract task-specific spatiotemporal features from the inherently noisy EEG signals, to assess the movement components. The leave-one-out evaluation results suggest that our CNN generalizes well across subjects, which is promising in being used for new patients, with minimal to no training. 
\par 
Since a successful motor recovery requires both repetitive movements~\cite{krebs2016beyond} and active participation~\cite{bernstein1966co}, an adaptation based on specific movement components could further tailor rehabilitation and promote plasticity. For example, an adaptive robotic therapy could target each one of the two movement components: for movement intent, the robot's assistance would adapt to encourage voluntary movements; for RT, the robot would use the percentage of fast RT as an indicator of functional motor recovery or even relate it to motor skill consolidation. In addition, while we focused here on rehabilitation of upper extremity, it can be argued that the two adaptation criteria apply to lower extremity robots as well. We have previously shown the similarity between the distribution of RTs in the upper and lower extremities~\cite{michmizos2014modeling}. Hence, our work can potentially target a variety of rehabilitation tasks. In that sense, our work can complement the current ``assist-as-needed'' controllers that rely on robot-derived metrics of motor performance for adaptation~\cite{krebs2003rehabilitation,michmizos2012assist,michmizos2015robot}, by adding an EEG-based measurement that may not be affected by the sensorimotor impairment. 

\section{Conclusion}
Overall, the paper discusses our ongoing efforts to decompose motor recovery into its fundamental components and, therefore, it has the potential to shape the design of new rehabilitation robots with a “built-in” adaptability. The integration of the two movement components, namely the intent to move and RT, into the adaptation of assistive devices will result in a) rehabilitation robots that deliver more personalized therapy, by targeting separate attributes of the activity-dependent neural plasticity, and b) an increase of the knowledge of how the brain learns. Overall, this work aims to help in the transition from the evidence-based to science-based rehabilitation, and optimize the collaborative interaction between humans in need and robots in service.
\bibliographystyle{IEEEtran}
\bibliography{ref}
\end{document}